%% file: main.tex
\documentclass[10pt,twocolumn,letterpaper]{article}
\usepackage[pagenumbers]{cvpr}
\input{preamble}

\def\paperID{-----}
\def\confName{CVPR}
\def\confYear{2026}

\title{V-DPM\@: 4D Video Reconstruction with Dynamic Point Maps}

\newcommand{\equalcontributor}{\@myfnsymbol{1}}

\author{Edgar Sucar$^\ast$ \hspace{1em} Eldar Insafutdinov$^\ast$ \hspace{1em} Zihang Lai \hspace{1em} Andrea Vedaldi\\
Visual Geometry Group (VGG), University of Oxford\\
{\tt\small \{edgarsucar,zlai,eldar,vedaldi\}@robots.ox.ac.uk}
}

\begin{document}
\twocolumn[{
\maketitle
\begin{center}
\includegraphics[width=\linewidth]{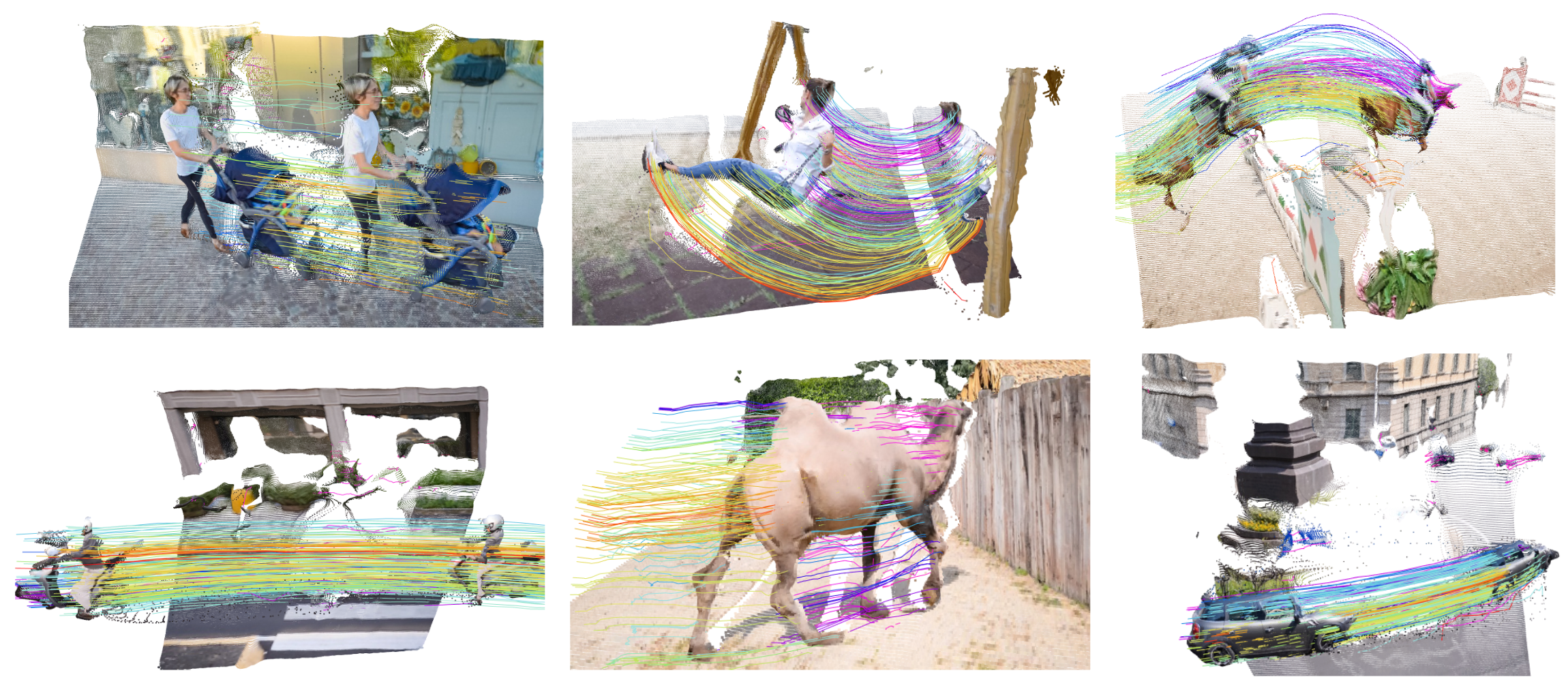}
\end{center}
\captionof{figure}{\textbf{\method results.}
We propose a method for extending state-of-the-art static 3D reconstructors like VGGT with Dynamic Point Maps (DPMs). Given a video snippet, \method reconstructs the 3D motion of the scene (i.e., the scene flow), along with its 3D shape and the camera parameters.
Because of DPMs, the same representation captures both the static background and complex non-rigid motion.}%
\label{fig:splash}
\vspace{2em}
}]
\def\thefootnote{*}\footnotetext{Equal contribution.} 

\input{sec/0_abstract}
\input{sec/1_intro}
\input{sec/2_related}
\input{sec/3_method}
\input{sec/4_experiments}

\input{sec/5_conclusions}

\paragraph*{Acknowledgements.}

We thank the ERC CoG 101001212-UNION\@.
The authors acknowledge the use of resources provided by the Isambard-AI National AI Research Resource (AIRR)~\cite{mcintosh2024isambard}.
Isambard-AI is operated by the University of Bristol and is funded by the UK Government’s Department for Science, Innovation and Technology (DSIT) via UK Research and Innovation; and the Science and Technology Facilities Council [ST/AIRR/I-A-I/1023].

{
\small
\bibliographystyle{ieeenat_fullname}
\bibliography{vedaldi_general,vedaldi_specific,main}
}

\input{sec/X_suppl}
\end{document}

%% file: preamble.tex
\usepackage[dvipsnames]{xcolor}
\usepackage{multirow}
\usepackage{xspace}

\definecolor{cvprblue}{rgb}{0.21,0.49,0.74}
\usepackage[pagebackref,breaklinks,colorlinks,allcolors=cvprblue]{hyperref}

\newcommand{\method}{V-DPM\xspace}

\makeatletter
\renewcommand{\paragraph}{%
  \@startsection{paragraph}{4}%
  {\z@}{-0.5em}{-0.5em}%
  {\normalfont\normalsize\bfseries}%
}
\makeatother

%% file: sec/0_abstract.tex
\newcommand{\duster}{DUSt3R\xspace}

\begin{abstract}
Powerful 3D representations such as \duster's invariant point maps, which encode 3D shape and camera parameters, have significantly advanced feed-forward 3D reconstruction.
While point maps assume static scenes, Dynamic Point Maps (DPMs) extend the concept to dynamic 3D content by also representing scene motion.
However, DPMs have so far been limited to image pairs and, like \duster, require post-processing via optimisation when more than two views are involved.
We argue that DPMs are more useful when applied to videos and introduce \method to demonstrate this.
First, we show how to set up DPMs for videos to optimise representational power, facilitate neural prediction, and enable reuse of pretrained models.
Second, we implement these ideas on top of VGGT, a recent powerful 3D reconstructor.
Although VGGT was trained on static scenes, we show that a modest amount of synthetic data suffices to adapt it into an effective \method predictor.
This yields state-of-the-art 3D and 4D reconstruction in dynamic settings.
In particular, unlike recent dynamic extensions of VGGT such as P3, DPMs recover not only dynamic depth but also the 3D motion of every point in the scene.
Code and demo are available at \url{https://www.robots.ox.ac.uk/~vgg/research/vdpm/}.
\end{abstract}

%% file: sec/1_intro.tex
\section{Introduction}%
\label{sec:intro}

We consider the problem of reconstructing dynamic 3D scenes from videos by means of feed-forward neural networks.
This class of models has progressed rapidly in the past few years, often driven by the introduction of powerful 3D representations.
Perhaps the best example is DUSt3R~\cite{wang24dust3r:}, which proposed \emph{viewpoint-invariant point maps}.
These representations encode both 3D shape and camera motion and are well suited to prediction by neural networks.
Point maps have since been used in many follow-up works.
A particularly important extension was the introduction of networks~\cite{tang25mv-dust3r:,wang25vggt,yang25fast3r:,keetha25mapanything:} that can process more than two views in a single feed-forward pass.
This has resulted in a new class of multi-view 3D reconstructors that are fast, robust, and accurate.

A significant limitation of point maps in their original formulation is that they do not support dynamic content.
This is important because, in most real-life applications—from entertainment to robotics—one must reconstruct \emph{dynamic} events in which objects move and deform over time.
Some follow-up works, like MonST3R~\cite{zhang24monst3r:} and others~\cite{wang25continuous,wang25p3} that tackle 4D reconstruction, either do not use point maps or—if they do—must pair them with additional components, such as 2D point trackers, to capture dynamic 3D information (e.g., scene flow).

Dynamic Point Maps (DPM)~\cite{sucar25dynamic} remove this limitation by extending point maps to account for scene motion.
The new representation achieves both \emph{viewpoint} and \emph{time} invariance, and can thus capture in a unified manner 3D shape, 3D motion, camera intrinsics, and camera motion.
However, the work of~\cite{sucar25dynamic} shares the same limitation as the original DUSt3R in that it only computes \emph{pairwise} DPMs; processing more than two images requires post-processing via optimisation methods.
A further question is how to best extend DPMs to multiple images: potentially there is a different point map for every combination of viewpoints and times in the input sequence, so the number of maps could grow quadratically with sequence length.

In this work, we propose and investigate \method, a multi-view (video) extension of DPMs.
We begin by proposing a design that extends recent multi-view feed-forward reconstruction architectures to support DPMs.
First, the backbone of the network is tasked with predicting time-varying point maps, one for each input image.
These point maps are viewpoint-invariant but time-varying, since we relax the static-scene assumption; nevertheless, the backbone is well suited to predict them.
We then add decoders that, given the signals computed by the backbone, output viewpoint- and time-invariant point maps.
These decoders effectively reconstruct the scene with respect to a fixed reference viewpoint (that of the `first' image) and an arbitrarily selected reference time.
In this way, all input images contribute to a reconstruction at a chosen viewpoint and time, pooling and fusing information from the inputs.
By varying the reference time, one can reconstruct the scene at any instant and recover scene flow.

This design has multiple advantages.
First, it conceptually splits the reconstruction task into two phases that build on each other effectively.
In the first phase, a viewpoint-invariant, time-varying reconstruction is performed.
In the second phase, additional layers analyse the phase-one outputs to establish time invariance, implicitly producing dynamic correspondences across the time-varying reconstructions.

Second, the backbone of the new model has the same architecture and similar statistics to the original static model.
This makes it easy to \emph{extend} an existing static model to support dynamic reconstruction, introducing DPMs gradually.
This allows fine-tuning an existing static reconstruction network instead of training a new model from scratch, which greatly reduces training cost and, in particular, the need for 4D annotated data.

We take advantage of this design by building \method on top of the pre-trained VGGT~\cite{wang25vggt} model.
With this, we obtain strong 4D reconstruction performance: on standard benchmarks, we more than halve the error rate compared to analogous feed-forward reconstructors such as DPM, MonST3R, and St4rTrack~\cite{feng25st4rtrack:}.
This is particularly notable because the original VGGT model was trained for static reconstruction only and had not seen any dynamic data prior to fine-tuning.
\method can effectively steer this model toward dynamic reconstruction.
See~\cref{fig:splash} for dynamic reconstruction results.

To summarise, our contributions are as follows.
First, we introduce a multi-image/video extension of DPMs.
Second, we show how this naturally leads to an extension of state-of-the-art multi-view feed-forward reconstructors.
Third, we show that, using this approach, a multi-view static 3D reconstruction network can be fine-tuned to achieve state-of-the-art 4D reconstruction with relatively little training data.

%% file: sec/2_related.tex
\section{Related Work}%
\label{sec:related-work}

\paragraph{Feed-forward static reconstruction.}

While machine learning and deep neural networks have long been used to assist 3D reconstruction from images, they were mostly employed alongside classical optimisation-based methods rooted in visual geometry, solving subtasks like feature matching and depth estimation.
More recently,
DUSt3R~\cite{wang24dust3r:} and its follow-up
MASt3R~\cite{duisterhof24mast3r-sfm:} introduced feed-forward models that, given an image pair, estimate 3D shape as well as camera intrinsics and extrinsics in a single pass.
These works demonstrated the usefulness of the viewpoint-invariant point map representation, which had already been partially recognised by Learning to Recover 3D Scene Shape~\cite{yin21learning} in the monocular setting.
Pow3R~\cite{jang25pow3r:} further added the ability to specify cameras instead of estimating them.

A shortcoming of DUSt3R and MASt3R is that they operate on image pairs only and require test-time optimisation to fuse additional views.
Subsequent works like
MV-DUST3R~\cite{tang25mv-dust3r:},
Fast3R~\cite{yang25fast3r:},
Flare~\cite{zhang25flare:},
MapAnything~\cite{keetha25mapanything:},
and VGGT~\cite{wang25vggt}
extended DUSt3R to multiple views.
VGGT, in particular, achieved better feed-forward performance than prior methods that rely on test-time optimisation.
CUT3R~\cite{wang25continuous} and Point3R~\cite{wu25point3r:} added incremental reconstruction, and $\pi^3$~\cite{wang25p3} further improved performance across the board.

\paragraph{Feed-forward dynamic reconstruction.}

DUSt3R was first directly extended to dynamic (4D) reconstruction in MonST3R~\cite{zhang24monst3r:}.
However, that formulation is insufficient to recover 4D motion intrinsically and must be paired with a 2D tracker to do so.
Dynamic Point Maps (DPMs)~\cite{sucar25dynamic} extend point maps to a viewpoint- and time-invariant representation.
They show that this representation is \emph{complete} in the sense that it can be used to recover all key 3D and 4D information about the scene, including scene flow.
St4RTrack~\cite{feng25st4rtrack:}, a concurrent work, proposes a related formulation.

Other feed-forward models perform partial dynamic reconstruction: they recover and align dynamic depth but do not recover scene motion without auxiliary components such as a 2D point tracker.
Examples include Align3R~\cite{lu25align3r:}, the aforementioned CUT3R and $\pi^3$, PAGE-4D~\cite{zhou25page-4d:}, and Geo4D~\cite{jiang25geo4d}, the latter building on video diffusion.

\paragraph{Other dynamic reconstruction approaches.}

Monocular dynamic 3D reconstruction has a long history, with earlier work by Bregler~\etal~\cite{bregler00recovering} and Torresani~\etal~\cite{torresani08nonrigid}.
One influential recent work is MegaSAM~\cite{li25megasam:}, which combines feed-forward predictors (for depth) with optimisation-based non-rigid reconstruction.

%% file: sec/3_method.tex
\section{Method}%
\label{sec:method}

\begin{figure}[b]
\centering
\includegraphics[width=\linewidth]{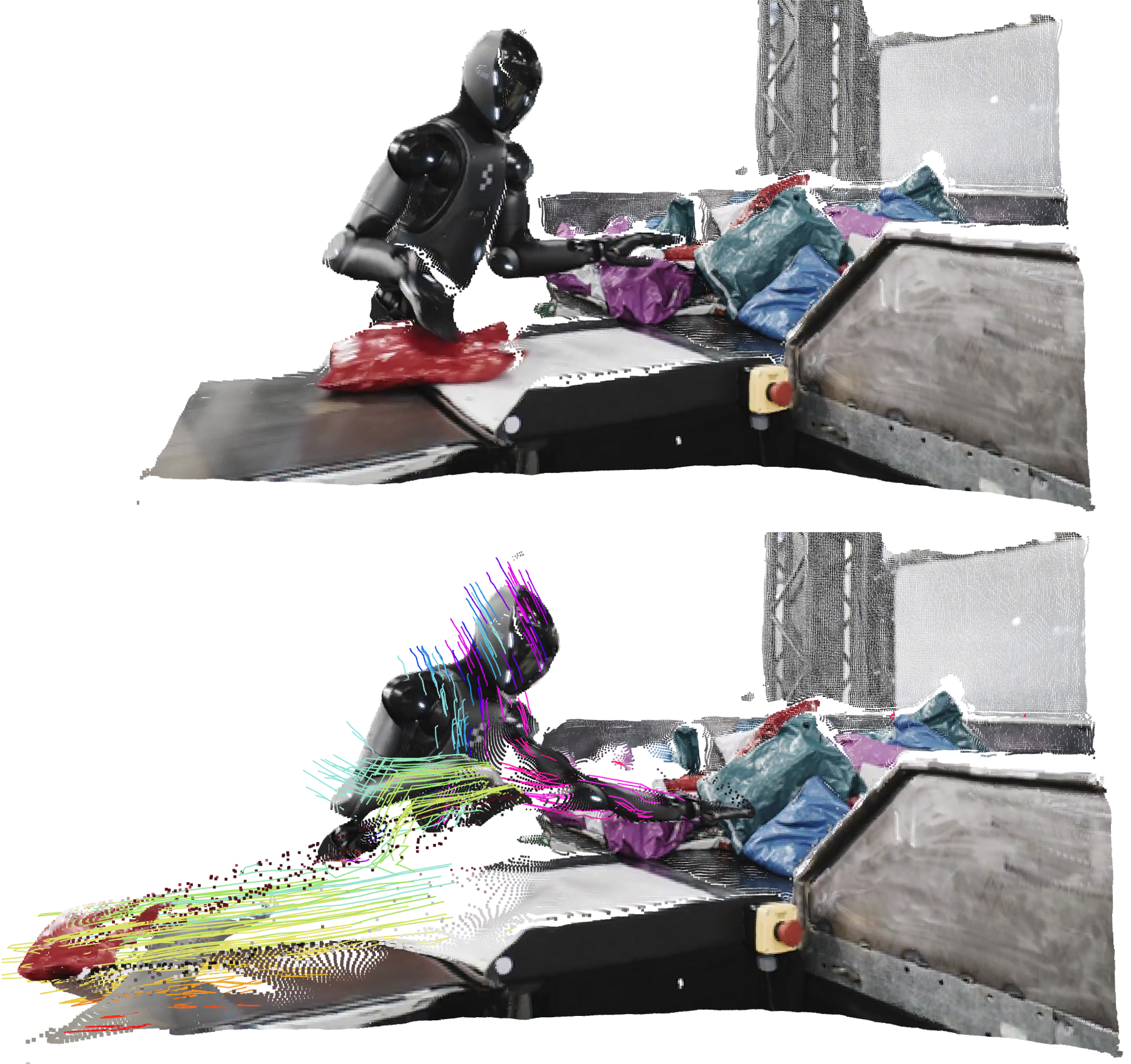}
\caption{Dynamic point maps of a robot doing a manipulation task.}%
\label{f:robot}
\end{figure}

We propose a multi-view extension of Dynamic Point Maps (DPMs)~\cite{sucar25dynamic} to represent and reconstruct dynamic 3D scenes from several images or a video, see~\cref{f:robot}.
We begin by reviewing DPMs in \cref{sec:dpm}.
Next, in \cref{sec:from-static-to-dynamic}, we describe our many-images extension.
Finally, in \cref{sec:implementation}, we describe a specific implementation built on top of the VGGT model.

\begin{figure*}[t]
\centering
\includegraphics[width=\linewidth]{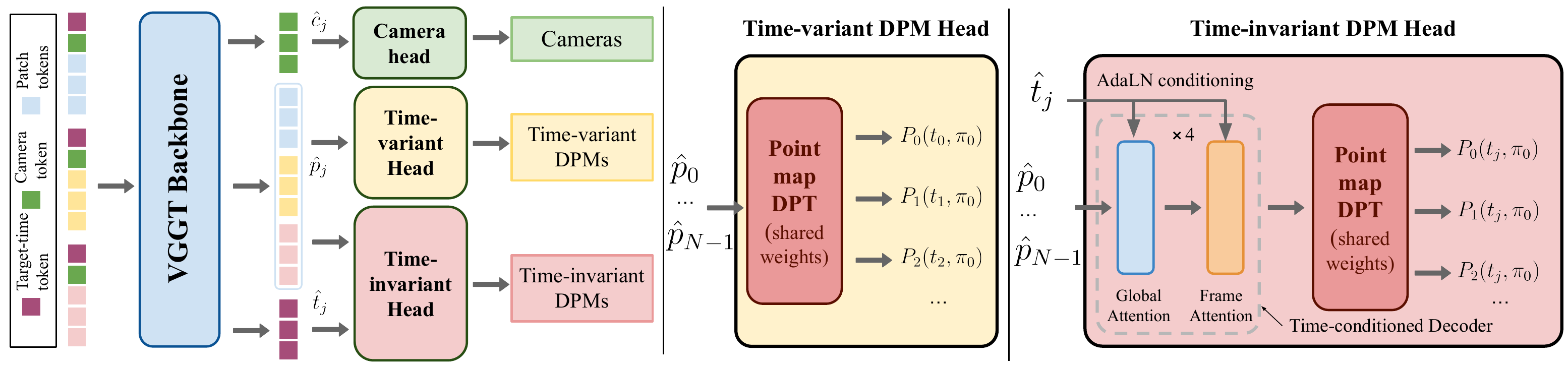}
\caption{
\textbf{Model architecture of \method.}
Our model decodes both time-variant point maps as in MonST3R~\cite{zhang24monst3r:} and time-invariant point maps corresponding to a fixed timestamp $t_j$ via the proposed time-conditioned decoder.
}%
\label{fig:diagram}
\vspace{1em}
\end{figure*}

\begin{figure}[t]
\centering
\! \! \! \includegraphics[width=0.7\linewidth]{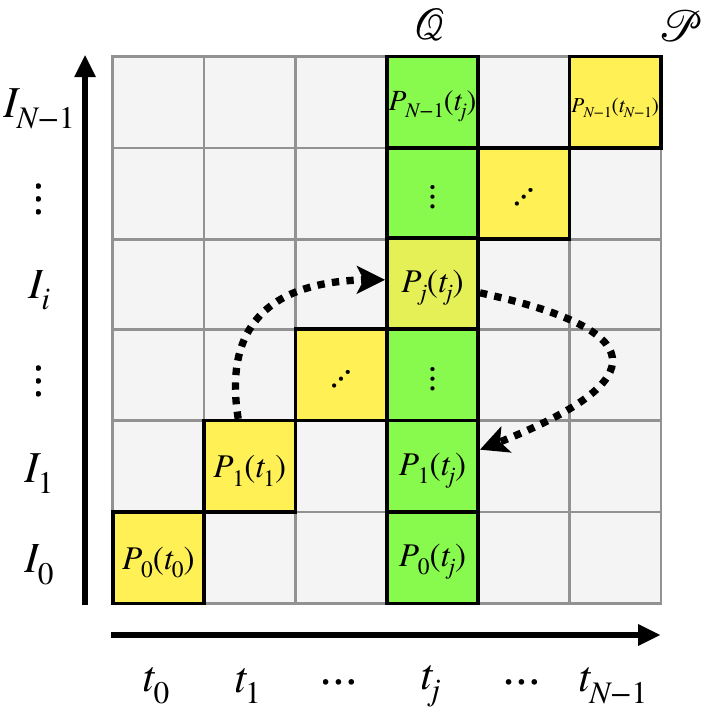}
\caption{
\textbf{\method point maps.}
The point maps $\mathcal{P}$ (yellow) are time-variant: they predict the 3D points at their respective input timestamps (we do not show the argument $\pi_0$ for compactness).
The point maps $\mathcal{Q}$ (green) are time-invariant: they predict the 3D points at a common reference timestamp $t_j$.
}%
\label{fig:point-map}
\end{figure}

\begin{figure}[t]
\centering
\includegraphics[width=0.93\linewidth]{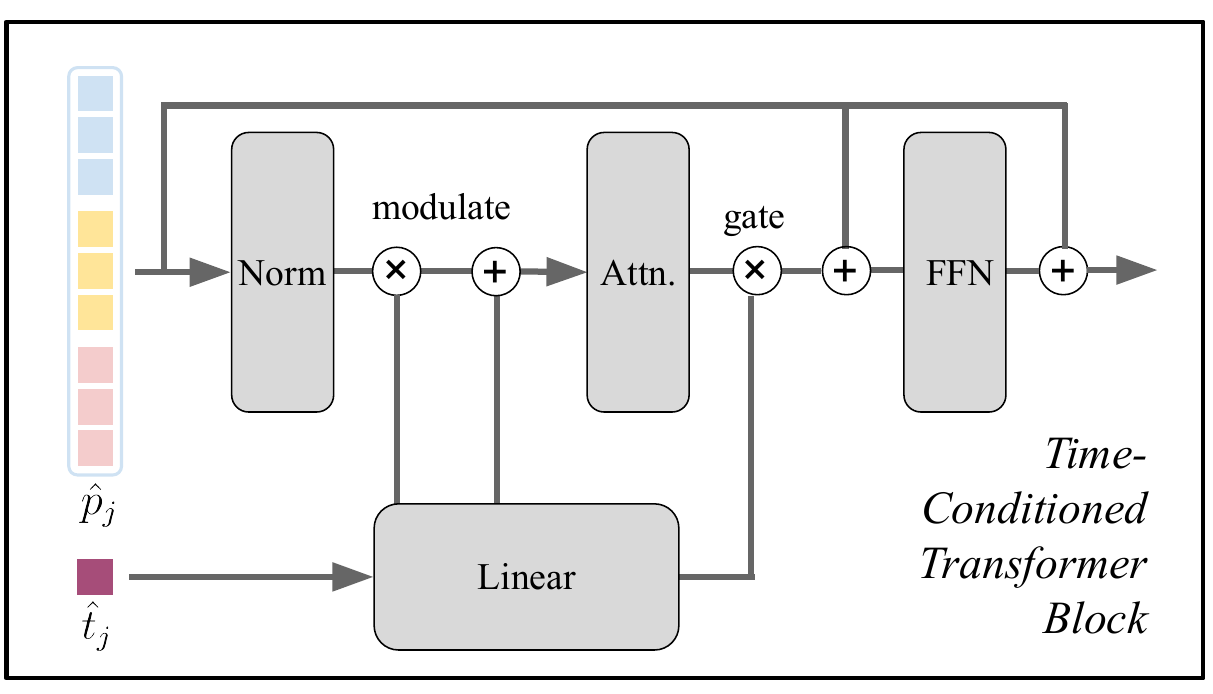}
\caption{
\textbf{Transformer block} in the time-conditioned decoder.
Conditioning is implemented via adaptive LayerNorm~\cite{perez18film:,peebles23scalable}.
}%
\label{fig:time_block}
\end{figure}

\subsection{Dynamic Point Maps}%
\label{sec:dpm}

Consider a sequence of images $I_i \in \mathbb{R}^{3\times H\times W}$ for $i=0,1,\dots,N-1$ and let $u \in \{0,\dots,H-1\} \times \{0,\dots,W-1\}$ denote a pixel location.
Denote by $t_i \in \mathbb{R}$ the \emph{timestamps} and by $\pi_i \in SE(3)$ the \emph{viewpoints} (camera extrinsics) associated to each image $I_i$.
Usually the images are video frames, but this is not strictly necessary because nothing in our design assumes a particular temporal ordering of the images: the timestamps $t_i$ can be thought of as image indices.

The \emph{Dynamic Point Map}~\cite{sucar25dynamic} representation $P$ associated to $I$ is a collection of point clouds
\begin{equation}\label{eq:dpm}
P_i(t_j,\pi_k) \in \mathbb{R}^{3\times H\times W}.
\end{equation}
These point clouds are in the form of images and associate a 3D point $P_i(t_j,\pi_k)(u)$ to each image pixel $u$.
Specifically, the index $i$ indicates that the 3D points in $P_i$ correspond to the pixels in image $I_i$.
The points are expressed relative to the specified viewpoint $\pi_k$, which, crucially, can differ from the viewpoint $\pi_i$ of the image $I_i$ itself.
Likewise, points are given at the position they occupy at time $t_j$, which can differ from the time $t_i$ of the image.

\paragraph{Pair-wise DPMs.}

The work of~\cite{sucar25dynamic} shows that, given two images $I_0$ and $I_1$, the four point maps
$
P_0(t_0,\pi_0)
$,
$
P_0(t_1,\pi_0)
$,
$
P_1(t_0,\pi_0)
$,
$
P_1(t_1,\pi_0)
$
encode all the information required to reconstruct the 3D shape and motion of the scene, as well as the camera intrinsics and camera motion, at least for the two given images.
For example, we can determine whether pixels $u$ and $v$ in images $I_0$ and $I_1$ correspond by checking if
$
P_0(t_0,\pi_0)(u) = P_1(t_0,\pi_0)(v).
$
This works because points are expressed relative to the same viewpoint $\pi_0$ and at the same time $t_0$.
The latter is key because it allows establishing a correspondence even if the point moves in 3D space between the two images.
The difference
$
P_0(t_1,\pi_0)(u) - P_0(t_0,\pi_0)(u)
$
gives instead the scene flow for pixel $u$ in image $I_0$.

The main drawback of this formulation is that it is limited to pairs of images.
If one has more than two images, then, like \duster, the network can be applied to pairs of them, but then post-processing via optimisation is needed to fuse the results, as done also in~\cite{sucar25dynamic}.
Below we discuss how to remove this limitation.

\paragraph{Comparison to static point maps.}

It is useful to note the difference compared to `static' point map representations like DUSt3R~\cite{wang24dust3r:}.
In this case, since the scene is static, there is no notion of time, and one predicts just two point maps
$
P_0(\pi_0) = P_0(t_0,\pi_0) = P_0(t_1,\pi_0)
$
and
$
P_1(\pi_0) = P_1(t_0,\pi_0) = P_1(t_1,\pi_0),
$
which makes it impossible to recover the dynamic quantities we expressed above.
However, this connection suggests that one may start from a pretrained model like DUSt3R and extend it to support DPMs with minimal changes and limited fine-tuning.
This is what the authors of~\cite{sucar25dynamic} did: they added new heads to the DUSt3R model to predict the four point maps above and fine-tuned the model using relatively simple 4D datasets like Kubric~\cite{greff22kubric:}.

\subsection{Multi-view DPMs}%
\label{sec:from-static-to-dynamic}

Next, we move to our multi-view extension of DPMs in pursuit of a neural network capable of feed-forward 4D reconstruction of a dynamic scene.
Note that \cref{eq:dpm} is not limited to pairs of images.
In fact, letting $i$, $j$, and $k$ vary in $\{0,\dots,N{-}1\}$ yields $N^3$ point maps.
Fortunately, these point maps are redundant.
By definition, point maps that differ only by viewpoint $\pi_k$ are related by a rigid transformation.
Hence, as long as we express all point maps relative to a \emph{common viewpoint} $\pi_0$ (achieving viewpoint invariance), the remaining point maps can be inferred once the cameras are recovered.
Thus, without loss of generality, we can limit ourselves to computing point maps for viewpoint $\pi_0$ only, reducing the total to $N^2$.
Even so, predicting $N^2$ point maps in a single feed-forward pass is computationally expensive; we therefore look for a useful subset.

Our idea is to consider two subsets of point maps, computed in sequence.
First, we task the neural network with predicting point maps (\cref{fig:point-map}, yellow)
\begin{equation}\label{eq:vdpm0}
\mathcal{P}
=(P_0(t_0,\pi_0),~
P_1(t_1,\pi_0),~
\dots,~
P_{N-1}(t_{N-1},\pi_0)).
\end{equation}
These point maps are indeed viewpoint invariant, in the sense that they share the same viewpoint $\pi_0$.
However, they are \emph{time-variant} (\cref{fig:diagram}), as each $P_i(t_i,\pi_0)$ uses the timestamp $t_i$ of image $I_i$.

Because they lack time invariance, these point maps cannot be used directly to reconstruct dynamic quantities like scene flow (\cref{sec:dpm}).
These point maps are similar to the ones computed by MonST3R~\cite{zhang24monst3r:} for pairs of images (as well as a subset of the ones computed by DPM and St4rTrack).
More importantly for us, these are similar to the point maps \emph{already} predicted by off-the-shelf models like VGGT\@.
Those in fact output point maps $P_i(\pi_0)$, one for each input image $I_i$.
For static scenes, these are identical to $P_i(\pi_0)$, so fine-tuning a pretrained model to output $P_i(t_i,\pi_0)$ is straightforward.

Once computed, point maps~\eqref{eq:vdpm0} reconstruct all scene points in the same reference frame $\pi_0$ where, up to scene motion, they line up.
From there, we  add network decoders to predict the point maps (\cref{fig:point-map}, green)
\begin{equation}\label{eq:vdpm1}
\mathcal{Q}
=
(
  P_0(t_j,\pi_0),~
  P_1(t_j,\pi_0),~
  \dots,~
  P_{N-1}(t_j,\pi_0)
),
\end{equation}
which, together with $\mathcal{P}$, amounts to $2N-1$ different point maps in a single feed-forward pass of the overall model.
These additional point maps~\eqref{eq:vdpm1} are the same as~\eqref{eq:vdpm0}, but expressed with respect to the same reference timestamp $t_j$ (\cref{fig:diagram}), thus achieving both viewpoint and time invariance.
This also decomposes the recovery of a viewpoint- and time-invariant representation into two logical steps, which, as we will see below, helps the network design.
Intuitively, as indicated by the arrows in \cref{fig:point-map}, to determine $P_1(t_j,\pi_0)$, i.e., the location of points $P_1$ at time $t_j$, the second stage of the network can match $P_1(t_1,\pi_0)$ to $P_i(t_j,\pi_0)$ (both computed in stage 1) to find out how the 3D points `move'.

There is a further benefit to this scheme.
Computing \cref{eq:vdpm1} amounts to reconstructing the full 3D scene for a specific timestamp $t_j$.
As we vary $t_j$, we obtain versions of the same scene at all timestamps by re-running only the decoder for \cref{eq:vdpm1}, reusing \cref{eq:vdpm0} and most backbone computations.
In fact, it is possible to reuse even more calculations by minimising the number of network layers that depend on the choice of $t_j$.

\subsection{Implementation}%
\label{sec:implementation}

Concretely, our goal is to implement a neural network that can, given $N$ images $I_0,\dots,I_{N-1}$ as input, predict both point maps~\eqref{eq:vdpm0} and~\eqref{eq:vdpm1}, i.e.,
$
(\mathcal{P}, \mathcal{Q}) = \Phi(I_0,\dots,I_{N-1}).
$
As discussed above, we want to leverage pretrained models for static scene reconstruction to minimise training time and data requirements, particularly due to the challenges of obtaining large-scale dynamic 4D datasets.

We build on VGGT~\cite{wang25vggt} as a pretrained backbone due to its excellent performance (even though it was never trained on dynamic data).
Recall that VGGT takes as input images $I_i$, $i=0,\dots,N{-}1$, and outputs cameras, per-image depth maps, and point maps.
For each input image $I_i$ it constructs image patch tokens $p_i$, a camera token $c_i$, and register tokens $r_i$; their concatenation $(p_i,c_i,r_i)$ is processed by an Alternating Attention Transformer to produce the output tokens $(\hat{p}_i,\hat{c}_i,\hat{r}_i)$.
We remove the redundant depth map prediction and fine-tune the rest of the network.
In VGGT the predicted tokens $\hat{p}_i$ are pulled from four layers of the backbone and decoded into point maps by a DPT head; we reuse this mechanism to predict the time-variant point maps~\eqref{eq:vdpm0} (yellow block in \cref{fig:diagram}).
Likewise, the original camera pose regressor is used as is to predict camera intrinsics and extrinsics from camera tokens $\hat{c}_i$.

\paragraph{Time-conditioned decoder.}

The key challenge is to compute the point maps~\eqref{eq:vdpm1} at a fixed time $t_j$: $P_0(t_j,\pi_0),\dots,P_{N-1}(t_j,\pi_0)$.
Unlike time-variant point maps, the target timestamp no longer corresponds to an input frame and must be supplied as an additional input.
We seek an architecture that can jointly reason about motion and align dynamic points across all frames to the common time $t_j$.
To that end, we add a time-conditioned transformer decoder (\cref{fig:time_block}) with alternating frame and global attention blocks.
The decoder processes the same backbone features $\hat{p}_i$ used by the DPT decoder for the time-variant point maps~\eqref{eq:vdpm0}.
Its blocks iteratively transform these features to align all frames to $P_j(t_j,\pi_0)$, whose features remain unchanged.
Because the DPT takes tokens from four layers in the backbone as input, we apply the decoder to each layer, concatenate the outputs, and feed them to the DPT head.

To inform the decoder of the target time $t_j$, we introduce two changes.
First, we augment VGGT's input tokens with a target-time token $t_j$ (reusing notation), transformed by the backbone into output tokens $\hat{t}_j$.
Second, we condition the decoder's transformer blocks via adaptive LayerNorm (adaLN), following FiLM~\cite{perez18film:} and DiT~\cite{peebles23scalable}.
We remove learned scale and shift parameters from LayerNorm and instead modulate normalised patch tokens with linear projections of the target-time token $\hat{t}_j$; the self-attention outputs are further gated by a second projection (\cref{fig:time_block}).
Decoder outputs are then passed to the point map DPT head which shares weights with the original, ensuring the feature distribution matches backbone outputs $\hat{p}_i$.

In practice, we run the VGGT backbone once to obtain $\hat{p}_i$ and then decode any $P_i(t_j,\pi_0)$ by evaluating only the decoder conditioned on the desired $\hat{t}_j$, which saves significant computation as $\hat{t}_j$ varies.

\paragraph{Training.}

We leverage priors learned during large-scale VGGT pretraining and fine-tune on a mixture of static and dynamic datasets:
ScanNet++~\cite{yeshwanth23scannet:} and
BlendedMVS~\cite{yao20blendedmvs:} for static scenes, and
Kubric-F~\cite{greff22kubric:},
Kubric-G~\cite{sucar25dynamic},
PointOdyssey~\cite{zheng23pointodyssey:}, and 
Waymo~\cite{sun20scalability} for dynamic data.
We process the training data following DPM, extending it to video snippets.
Differently from DPM, we scale ground-truth point maps to have unit mean distance to the origin, and let the network predict the correct scale as in VGGT training.
During training, we sample video snippets of 5, 9, or 19 frames from the dataset; longer training samples ensure better generalisation to complex motions.
We supervise V-DPM with the confidence-calibrated loss from DPM plus camera pose regression as in VGGT\@.
Further training hyper-parameters are detailed in the Appendix.

%% file: sec/4_experiments.tex
\section{Experiments}%
\label{sec:experiments}

Our evaluation includes several benchmarks for 3D and 4D reconstruction.
In \cref{sec:exp-4d-reconstruction} we evaluate \method on dynamic 3D reconstruction tasks, and in \cref{sec:exp-video-depth} on (dynamic) depth prediction and camera pose estimation.

\begin{figure}[htb]
\centering
\includegraphics[width=\linewidth]{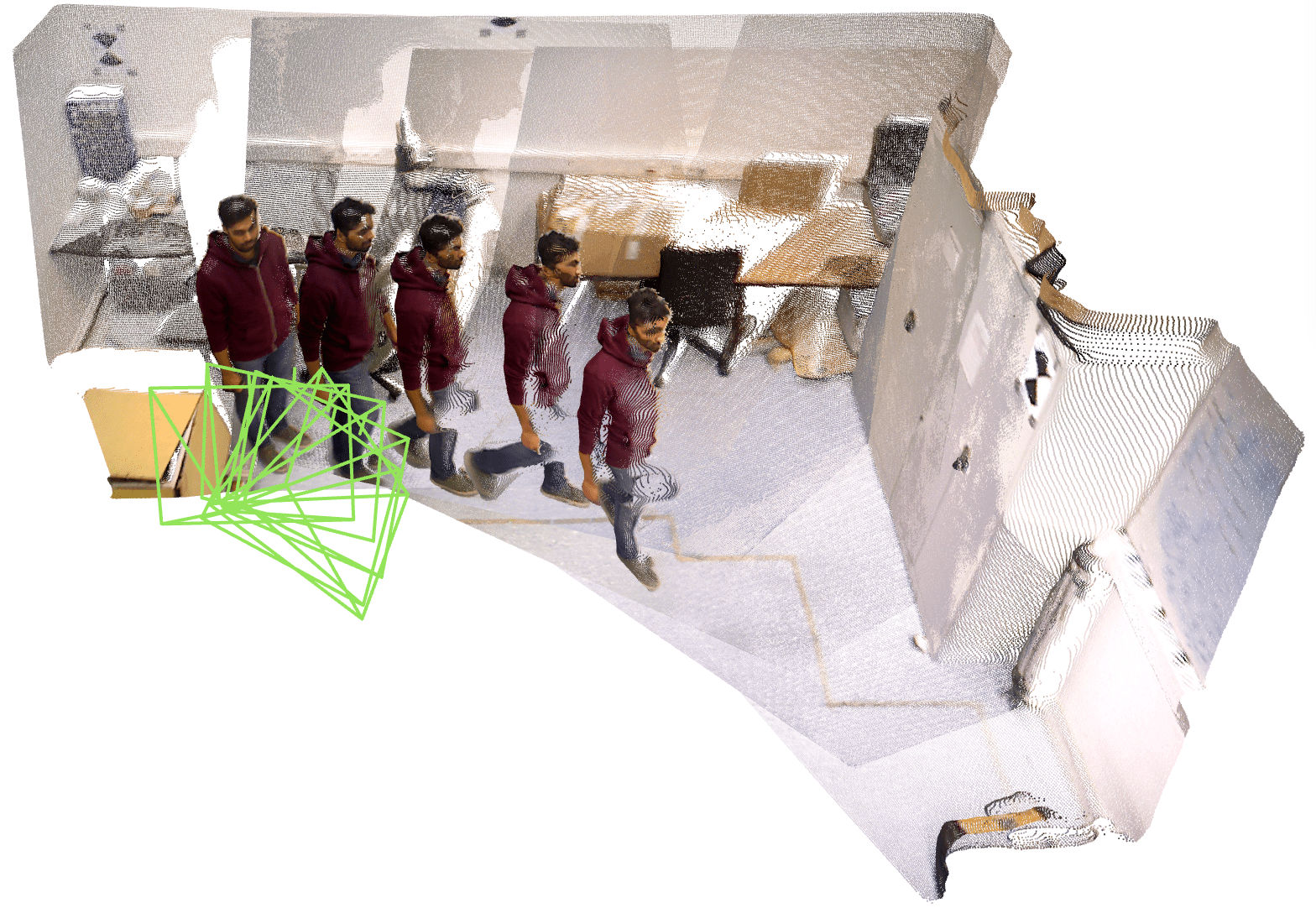}
\caption{Result of optimisation used for video depth and camera pose evaluation on a sequence from the Bonn dataset.}%
\label{f:bonn}
\end{figure}

\begin{figure*}[htb]
\centering
\includegraphics[width=\linewidth]{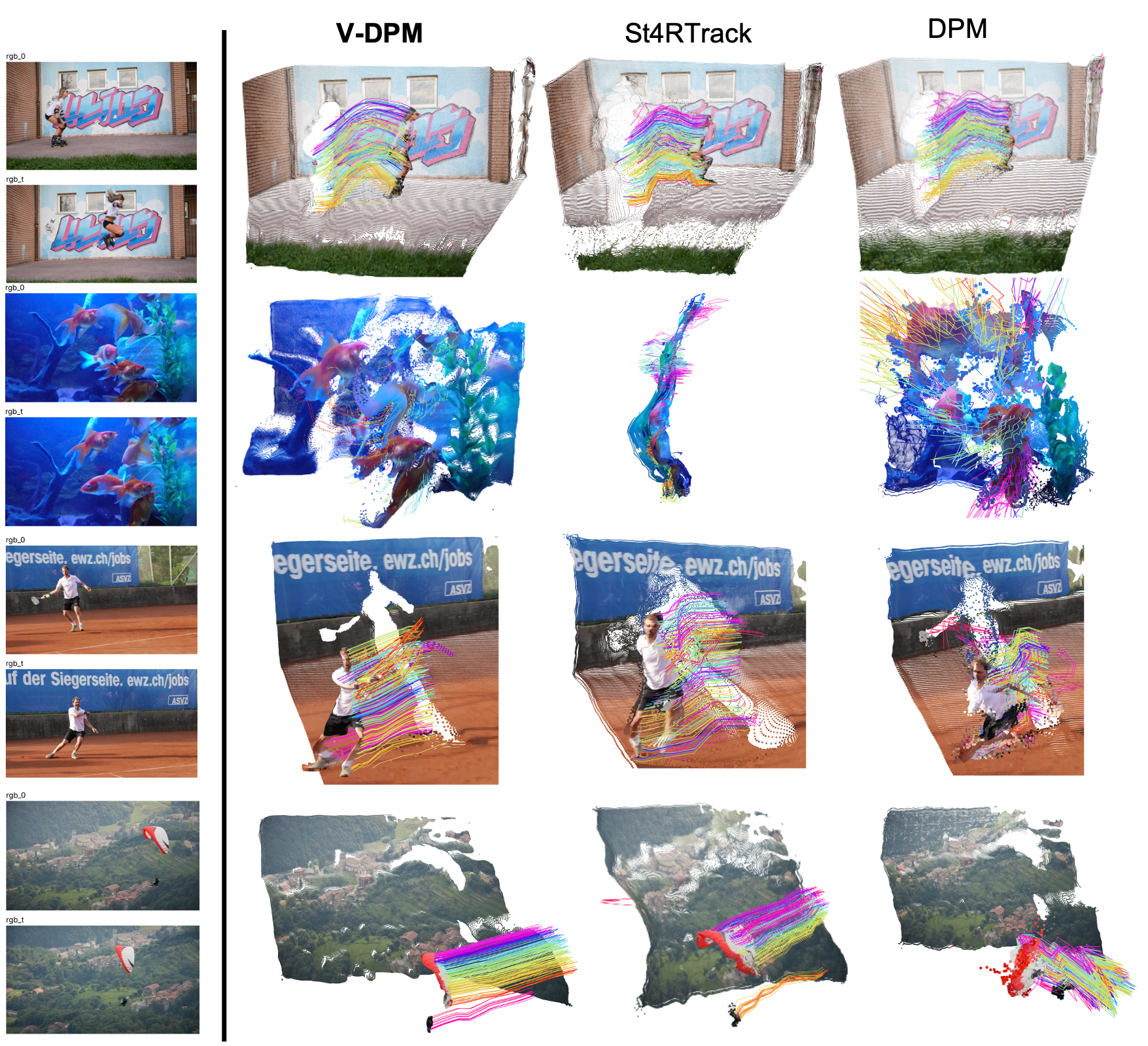}
\caption{Qualitative comparison of dynamic 3D tracking on the DAVIS dataset~\citep{perazzi2016benchmark}; results are reconstructed from 10-frame snippets. 
On the left we visualise the first and last input frames, and on the right we show the reconstructed point map $P_0(t_9, \pi_0)$ for the final timestep, as well as point trajectories over the entire snippet.
\method produces more accurate 3D reconstruction of the static scene background and generates smoother, more self-consistent 3D trajectories for the dynamic portions of the scene.}%
\label{f:qualitative-davis}
\end{figure*}

\subsection{4D Reconstruction}%
\label{sec:exp-4d-reconstruction}

\begin{table*}[htb]
\centering
\footnotesize
\setlength{\tabcolsep}{2pt}
\renewcommand{\arraystretch}{1.15}
\begin{tabular}{llcccccccccccccccc}
\toprule
\multirow{2}{*}{\textbf{Method}} &
\multicolumn{4}{c}{\textbf{PointOdyssey}} &
\multicolumn{4}{c}{\textbf{Kubric-F}} &
\multicolumn{4}{c}{\textbf{Kubric-G}} &
\multicolumn{4}{c}{\textbf{Waymo}} \\
\cmidrule(lr){2-5}\cmidrule(lr){6-9}\cmidrule(lr){10-13}\cmidrule(lr){14-17}
& $P_0(t_0)$ & $P_0(t_1)$ & $P_1(t_0)$ & $P_1(t_1)$ & $P_0(t_0)$ & $P_0(t_1)$ & $P_1(t_0)$ & $P_1(t_1)$ & $P_0(t_0)$ & $P_0(t_1)$ & $P_1(t_0)$ & $P_1(t_1)$ & $P_0(t_0)$ & $P_0(t_1)$ & $P_1(t_0)$ & $P_1(t_1)$ \\
\midrule

\multicolumn{17}{c}{\textbf{Margin: 2}} \\
\midrule
St4RTrack & --- & 0.145 & --- & 0.150 & --- & 0.149 & --- & 0.045 & --- & 0.173 & --- & 0.091 & --- & 0.228 & --- & 0.225 \\
TraceAnything& 0.159 & 0.159 & 0.163 & 0.163 & 0.069 & 0.071 & 0.071 & 0.070 & 0.086 & 0.087 & 0.088 & 0.087 & 0.151 & 0.151 & 0.148 & 0.148 \\
DPM  & 0.115 & 0.114 & 0.115 & 0.117 & 0.032 & 0.033 & 0.032 & 0.032 & 0.039 & 0.040 & 0.041 & 0.040 & 0.085 & 0.083 & 0.082 & 0.084 \\
\textbf{\method} & \textbf{0.029} & \textbf{0.030} & \textbf{0.032} & \textbf{0.032} & \textbf{0.018} & \textbf{0.019} & \textbf{0.018} & \textbf{0.018} & \textbf{0.023} & \textbf{0.024} & \textbf{0.024} & \textbf{0.023} & \textbf{0.064} & \textbf{0.064} & \textbf{0.064} & \textbf{0.064} \\
\midrule

\multicolumn{17}{c}{\textbf{Margin: 8}} \\
\midrule
St4RTrack & --- & 0.143 & --- & 0.146 & --- & 0.163 & --- & 0.059 & --- & 0.193 & --- & 0.113 & --- & 0.232 & --- & 0.261 \\
TraceAnything& 0.151 & 0.156 & 0.166 & 0.165 & 0.082 & 0.115 & 0.127 & 0.091 & 0.094 & 0.139 & 0.154 & 0.130 & 0.188 & 0.192 & 0.235 & 0.235 \\
DPM  & 0.101 & 0.103 & 0.103 & 0.104 & 0.030 & 0.050 & 0.044 & 0.039 & 0.041 & 0.068 & 0.065 & 0.051 & 0.085 & 0.085 & 0.083 & 0.084 \\
\textbf{\method} & \textbf{0.029} & \textbf{0.031} & \textbf{0.032} & \textbf{0.030} & \textbf{0.017} & \textbf{0.039} & \textbf{0.033} & \textbf{0.025} & \textbf{0.022} & \textbf{0.049} & \textbf{0.045} & \textbf{0.029} & \textbf{0.065} & \textbf{0.067} & \textbf{0.065} & \textbf{0.064} \\
\bottomrule
\end{tabular}
\caption{\textbf{2-View EPE} error for 4D reconstruction, reported for four point clouds (one for each image and time frame).}%
\label{tab:results_final_caps_bold}
\end{table*}

\begin{table}[ht]
\centering
\footnotesize
\setlength{\tabcolsep}{2pt}
\renewcommand{\arraystretch}{1.15}
\begin{tabular}{lcccc}
\toprule
\textbf{Method} & \textbf{PointOdyssey} & \textbf{Kubric-F} & \textbf{Kubric-G} & \textbf{Waymo} \\
\midrule
St4RTrack           & 0.137 & 0.153 & 0.201 & 0.167 \\
TraceAnything       & 0.152 & 0.107 & 0.126 & 0.119 \\
DPM                 & 0.114 & 0.088 & 0.109 & 0.103 \\
\textbf{\method}      & \textbf{0.032} & \textbf{0.027} & \textbf{0.035} & \textbf{0.042} \\
\bottomrule
\end{tabular}
\caption{\textbf{Tracking EPE} error reported for 10-frame snippets, evaluating dense tracks of all pixels in the first frame.}%
\label{tab:tracking}
\end{table}

First, we evaluate our model on the task of dynamic 3D reconstruction.
To make the model directly comparable to prior works like DPM~\citep{sucar25dynamic}, we assume first that there are two input views.
We use the DPM configuration of four datasets: PointOdyssey, Kubric-F, Kubric-G, and Waymo.
We randomly sample two views from the video either $2$ or $8$ frames apart.
The results in \cref{tab:results_final_caps_bold} report the End-Point Error on four predicted point maps $P_0(t_0, \pi_0)$, $P_0(t_1, \pi_0)$, $P_1(t_0, \pi_0)$ and $P_1(t_1, \pi_0)$.
In the table, we omit the symbol $\pi_0$ for brevity.
We only consider points for which there is valid 3D ground truth and normalise both predicted and ground-truth point maps to have unit mean norm.
Importantly, we evaluate reconstructions in the world coordinate frame defined by the first view $\pi_0$ (rather than the local camera frame for each view), so that the metric implicitly measures the accuracy of camera estimation and point tracking.
We compare our method with recent dense dynamic 3D reconstruction approaches: DPM~\citep{sucar25dynamic}, St4RTrack~\citep{feng25st4rtrack:} and TraceAnything~\citep{liu25trace}.
DPM and St4RTrack train on Kubric and PointOdyssey datasets, whereas TraceAnything proposes its own synthetic data engine for training.
\method convincingly outperforms prior work on all four benchmarks.
While St4RTrack and TraceAnything trade places on PointOdyssey and Kubric, our model achieves $\sim5\times$ lower error than both methods.

The experiment above primarily shows the effectiveness of our strategy for building \method on top of VGGT, as well as the ability of that model, which was trained on static data, to generalise to dynamic scenes with comparatively modest fine-tuning.
However, this evaluation does not assess the full potential of \method, which can process an entire video snippet at once.

Next, we consider a 3D dense tracking scenario, where we sample a video snippet of 10 frames, each spaced 2 frames apart.
We track 3D points in the first frame by computing the sequence $P_0(t_0, \pi_0), P_0(t_1, \pi_0), \ldots, P_0(t_9, \pi_0)$ and report an average EPE evaluated identically to the preceding experiment.
In the video setting (\cref{tab:tracking}), the original DPM's accuracy drops significantly compared to the 2-view reconstruction with 8 frames apart, since it can only make predictions on pairs of frames and cannot leverage temporal context.
Instead, \method maintains performance similar to the 2-view experiment owing to its capability to reason about temporal dynamics over the whole video snippet.

\paragraph{Qualitative comparison.}

In~\cref{f:qualitative-davis} we provide visualisations of 4D reconstructions of 10-frame snippets by \method, St4RTrack, and DPM\@.
\method produces smoother and more coherent motion trajectories, and is more robust, avoiding failure cases of previous methods.
For example, both DPM and St4RTrack fail on the fishtank sequence, and only \method plausibly reconstructs the human body pose of a tennis player for the end frame of the snippet (we visualise $P_0(t_9, \pi_0)$, which provides, for every pixel in image $I_0$, its final 3D position at time $t_9$).

\subsection{Video Depth and Camera Pose}%
\label{sec:exp-video-depth}

In this section, we evaluate the accuracy of joint dense reconstruction and pose estimation by our model.
With our hardware, we could only fine-tune \method for snippets of up to 20 frames (although we found it generalises to about 50 frames at test time).
To evaluate on longer sequences of hundreds of frames, we operate in a sliding-window manner and use a bundle-adjustment optimisation scheme similar to DUSt3R~\citep{wang24dust3r:,zhang24monst3r:} to fuse the windows.
The inputs to the optimisation are \method point map predictions computed on overlapping windows of frames; instead of pairwise constraints used in two-view methods, we use window constraints, as \method makes predictions over video snippets.
See~\cref{f:bonn} for an example result.

\begin{table}[ht]
\centering
\footnotesize
\setlength{\tabcolsep}{1pt}
\begin{tabular}{lc cc cc}
\toprule
\multirow{2}{*}{Category} & \multirow{2}{*}{Method} & \multicolumn{2}{c}{\textbf{Sintel}} & \multicolumn{2}{c}{\textbf{Bonn}} \\
\cmidrule(lr){3-4} \cmidrule(lr){5-6}
 &  & Abs Rel $\downarrow$ & $\delta<1.25$ $\uparrow$ & Abs Rel $\downarrow$ & $\delta<1.25$ $\uparrow$ \\
\midrule
\multirow{2}{*}{\footnotesize 1-frame} & Marigold & 0.532 & 51.5 & \textbf{0.091} & \textbf{93.1} \\
 & DepthAnythingV2 & \textbf{0.367} & \textbf{55.4} & 0.106 & 92.1 \\
\cmidrule(lr){1-6}
\multirow{3}{*}{\footnotesize \shortstack[l]{Video\\depth}} & NVDS & 0.408 & 48.3 & 0.167 & 76.6 \\
 & ChronoDepth & 0.687 & 48.6 & 0.100 & 91.1 \\
 & DepthCrafter & \textbf{0.292} & \textbf{69.7} & \textbf{0.075} & \textbf{97.1} \\
\cmidrule(lr){1-6}
\multirow{5}{*}{\footnotesize \shortstack[l]{Joint\\D\&P}} & Robust-CVD & 0.703 & 47.8 & --- & --- \\
 & CasualSAM & 0.387 & 54.7 & 0.169 & 73.7 \\
 & MonST3R & 0.335 & 58.5 & 0.063 & 96.4 \\
 & DPM & 0.311 & 58.0 & 0.064 & 94.8 \\
 & $\pi^3$ & \textbf{0.210} & \textbf{72.6} & \textbf{0.043} & \textbf{97.5} \\
 & \textbf{\method} & \underline{0.247} & \underline{69.4} & \underline{0.057} & \underline{97.3} \\ 
\bottomrule
\end{tabular}
\caption{\textbf{Video Depth Evaluation} on the Sintel and Bonn datasets.}%
\label{tab:depth_estimation_comparison}
\end{table}

\begin{table}[ht]
\centering
\footnotesize
\setlength{\tabcolsep}{1pt}
\begin{tabular}{lcccccc}
\toprule
\multirow{2}{*}{\textbf{Method}} & \multicolumn{3}{c}{\textbf{Sintel}} & \multicolumn{3}{c}{\textbf{TUM-dynamics}} \\
\cmidrule(lr){2-4} \cmidrule(lr){5-7}
 & ATE $\downarrow$ & RPE trans $\downarrow$ & RPE rot $\downarrow$ & ATE $\downarrow$ & RPE trans $\downarrow$ & RPE rot $\downarrow$ \\
\midrule
Robust-CVD & 0.360 & 0.154 & 3.443 & 0.189 & 0.071 & 3.681 \\
CasualSAM & 0.141 & \textbf{0.035} & \underline{0.615} & \underline{0.045} & 0.020 & 0.841 \\
DUST3R & 0.417 & 0.250 & 5.796 & 0.127 & 0.062 & 3.099 \\
MonST3R & 0.108 & 0.042 & 0.732 & 0.074 & 0.019 & 0.905 \\
DPM & --- & --- & --- & 0.056 & \underline{0.014} & 0.836 \\
$\pi^{3}$ & \textbf{0.074} & \underline{0.040} & \textbf{0.282} & \textbf{0.014} & \textbf{0.009} & \textbf{0.312} \\
\textbf{\method} & \underline{0.105} & 0.048 & 0.67 & 0.057 & 0.017 & \underline{0.34} \\
\bottomrule
\end{tabular}
\caption{Comparison of pose metrics on the Sintel and TUM-dynamics datasets.}%
\label{tab:pose}
\end{table}

\paragraph{Video-depth estimation.}

We report our results on the Sintel~\citep{butler12a-naturalistic} and Bonn~\citep{palazzolo2019iros} datasets.
This benchmark does not showcase the full capability of \method, which can track every pixel in every frame, and only evaluates the accuracy of time-variant point map~\eqref{eq:vdpm0} reconstruction.
The goal here is to show that our model is competitive with existing dynamic 3D reconstruction methods.
In~\cref{tab:depth_estimation_comparison}, we show that \method outperforms all prior art by a substantial margin except for a concurrent work, $\pi^3$~\citep{wang25p3}; however, this is likely an issue of scale, as they could train their model on 14 public datasets plus an internal dynamic dataset, whereas we only use 6.
$\pi^3$ is also stronger than our backbone VGGT\@.
In practice, their model is similar to VGGT, and we could integrate \method on top of their network to add motion reconstruction capabilities.

\paragraph{Camera pose estimation.}

We show results on camera pose estimation on Sintel and TUM-dynamics datasets in~\cref{tab:pose}.
Following MonST3R, we report Average Translation Error (ATE), Relative Translation Error (RPE trans), and Relative Rotation Error (RPE rot).
Similarly, V-DPM demonstrates competitive performance, and is only outperformed by $\pi^3$, which also outperforms our VGGT backbone on this task.
We expect that scaling up our training data and adopting a stronger, more recent backbone will close this gap.

%% file: sec/5_conclusions.tex
\section{Conclusions}%
\label{sec:conclusions}

We have presented \method, an extension of Dynamic Point Maps that supports one-shot 4D reconstruction from multi-frame monocular videos.
We have shown that this representation can be integrated into off-the-shelf 3D reconstruction networks in a natural and effective manner.
In particular, we take VGGT, a network trained to reconstruct static scenes, and extend it to a 4D video reconstructor using only a modest amount of compute and synthetic data.

The resulting model predicts time- and viewpoint-invariant 3D point clouds for each image.
Thus, it can be used to recover point motion (dense tracking) or to fuse point clouds extracted from different images captured at different times, effectively undoing deformations in the scene.
We show empirically that this model generalises well to diverse and challenging video snippets.
On scene motion reconstruction, it outperforms all previous feed-forward models by a large margin.
On static 3D and camera reconstruction, it is outperformed only by $\pi^3$, likely due to differences in training scale and backbone.
Overall, our training recipe highlights the potential of combining large datasets of static scenes---easy to obtain and auto-annotate---with a much smaller amount of synthetic data with accurate 4D annotations.
By using the \method representation, it is possible to learn effectively and seamlessly from both data sources.

One limitation of our evaluation is its scale, which is constrained by available resources.
Even so, our experiments highlight the potential of \method as a template for future 4D reconstructors and for applications such as VFX, video generation, world modelling, and vision-based control.

%% file: sec/X_suppl.tex
\clearpage
\setcounter{page}{1}
\maketitlesupplementary{}

\section{Training details}%
\label{sec:training-details}

Each training batch contains windows of frames randomly sampled from our dataset mixture.
We choose the central frame in the sampled snippet as the reference view that defines the coordinate system for multi-view reconstruction with the VGGT backbone.
As in VGGT, we randomise the length of the video snippet during training, which we found helps reconstruct longer and more complex motions.
Specifically, for each batch we sample a 5-, 9-, 13-, or 19-frame window.
To utilise the hardware more efficiently, we dynamically select the batch size depending on the snippet length: a window of length 5 allows for a batch size of $4$, whereas a 19-frame snippet can fit in VRAM only with a batch size of $1$.

We train our final model on 16 GH200 GPUs for 60 epochs.
During each epoch, we sample the following number of examples from each dataset: 5000 from Kubric-G, 5000 from Kubric-F, 15000 from PointOdyssey, 2500 from Waymo, 2500 from ScanNet++, and 2500 from BlendedMVS\@.
We use the AdamW optimiser with a base learning rate of $1.5 \times 10^{-4}$ and a cosine decay schedule.

Our dynamic point map reconstruction loss is defined for each pixel in each frame of every video snippet in the batch.
Naively averaging the loss across all valid pixels (i.e., those for which we have annotations) can lead to problems.
In particular, datasets with 4D annotations such as PointOdyssey often contain only sparse ground-truth 3D point tracks.
When averaging the loss across all points in the batch, the numerous annotated points from static 3D datasets can easily dominate the sparse dynamic 3D annotations from the synthetic training set.
As a result, the parts of the neural network responsible for dynamic reconstruction receive relatively small gradient updates.
To mitigate this, we propose the following normalisation scheme: we first average the loss within each example and then compute the average across the batch dimension.
This ensures that the magnitude of the loss is comparable across training samples.
We found this improves the accuracy of dynamic reconstruction.

\section{Network design ablation}%
\label{sec:ablation}

We train a smaller run of 35 epochs to test different design choices for the network architecture.
We compare four variants of the network design: (i) \textit{Original}, (ii) \textit{Decoder depth 2}, (iii) \textit{Addition conditioning}, and (iv) \textit{DPT decoder}.
The \textit{Original} is our complete model with four transformer blocks for decoding time-invariant point maps.
In \textit{Decoder depth 2}, we reduce the number of transformer blocks to two.
In \textit{Addition conditioning}, instead of using adaLN for time conditioning, we add the time token to the input tokens.
In \textit{DPT decoder}, we use no extra transformer layers for time-invariant decoding; instead, we make a copy of the DPT head and condition it directly through adaLN\@.

We evaluate the dynamic point map reconstruction on two views with a margin of 8 on the Kubric-G dataset; see \cref{sec:exp-4d-reconstruction}. The results verify the importance of each design element for the full performance of the model.

\begin{table}[h]
\centering
\begin{tabular}{lcc}
\toprule
 & \textbf{$P_0(t_1)$} & \textbf{$P_1(t_0)$} \\
\midrule
\method: Original & \textbf{0.0500} & \textbf{0.0472} \\
\method: Decoder depth 2 & 0.0518 & 0.0476 \\
\method: Addition conditioning & 0.0524 & 0.0484 \\
\method: DPT decoder & 0.0538 & 0.0502 \\
\bottomrule
\end{tabular}
\end{table}